%% file: iclr2027_conference.tex
\documentclass{article} % For LaTeX2e
\usepackage{iclr2027_conference,times}

\input{math_commands.tex}

\usepackage{hyperref}
\usepackage{url}
\usepackage{wrapfig}
\usepackage{graphicx}
\usepackage{booktabs}
\usepackage{multirow}
\usepackage{amssymb}
\usepackage[most]{tcolorbox}

\title{Presence Is Not Faithfulness: Figurative Vehicle Intrusion in Text-to-Image Generation}

\author{
Xiaoyu Ma \\
Southeast University \\
\texttt{xiaoyuma@seu.edu.cn}
\And
Chen Yang \\
Southeast University \\
\And
Hao Chen \\
Southeast University \\
\texttt{haochen303@seu.edu.cn}
}

\iclrfinalcopy % Uncomment for camera-ready version, but NOT for submission.
\begin{document}

\maketitle

\begin{abstract}
Text-to-image (TTI) models increasingly generate high-quality images from natural-language prompts, yet figurative language exposes a failure: a vehicle that should guide the depiction of a tenor may instead be rendered as a visible object.
We call this failure \textbf{Figurative Vehicle Intrusion}: the intruding content is textually licensed, but it is assigned the wrong visual role, showing that visual presence is not always faithfulness and that presence-oriented evaluation can miss such errors.
To study it systematically, we introduce \textbf{V}ehicle \textbf{I}ntrusion and \textbf{S}emantic \textbf{T}enor \textbf{A}ssessment (\textbf{VISTA}), a multilingual benchmark of figurative prompts organized by Figurative Form and Mapping Mechanism.
We further propose \textbf{V-Score}, a diagnostic question-answering metric that evaluates role-aware figurative faithfulness in generated images.
Evaluations on recent high-performing TTI models show that vehicle intrusion persists across languages and figurative categories.
As a lightweight mitigation, we introduce \textbf{VISTA-Guard}, which partially reduces vehicle intrusion and suggests a practical path toward more figuratively faithful TTI generation.
All resources will be released publicly.
\end{abstract}

\section{Introduction}
\begin{wrapfigure}{r}{0.38\textwidth}
    \centering
    \vspace{-2.2em}
    \includegraphics[width=0.96\linewidth]{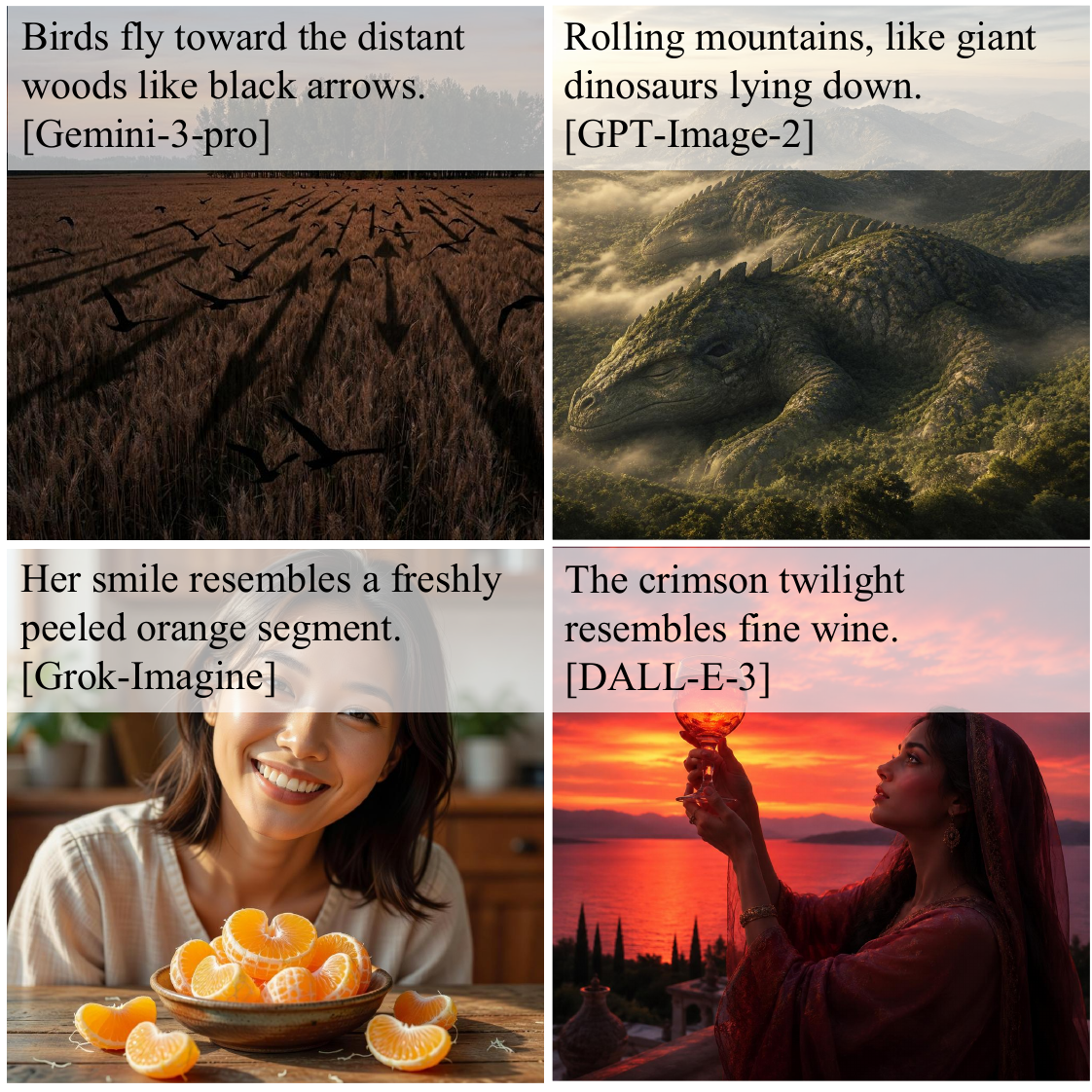}
    \caption{Figurative vehicles literalized by models.}
    \label{fig:intro-case}
    \vspace{-1.8em}
\end{wrapfigure}
\textbf{\textit{Should Generative Models Draw Every Word in Prompts?}}

Modern text-to-image (TTI) models, such as GPT-Image \citep{gptimage2} and Nano Banana Pro \citep{nanobananapro}, can now generate high-quality images that closely follow natural-language prompts, grounding objects and relations with impressive fidelity. 
Yet this strength also exposes a subtle failure when language is figurative: models may mistake a descriptive comparison for a drawable entity. 
As shown in Fig.~\ref{fig:intro-case}, birds described as ``black arrows'' may become arrows, while twilight likened to wine may bring wine into the scene. 
These cases reveal that \textbf{visual presence is not necessarily faithfulness}. 
The model does not fail by omitting prompt content, but by placing it in the \textbf{wrong visual role}: a figurative vehicle is rendered as a scene entity rather than used to shape the intended depiction.

This failure matters because visual-generation workflows are increasingly moving beyond manually cleaned prompts and operating directly on source prose.
In applications such as illustration, storyboarding, and narrative content creation, users often expect models to turn original passages into visual scenes \citep{storydiffusion,shahmohammadi2023vipe}.
For example, Fanqie Novel \citep{fanqie} supports illustration generation from selected novel passages, while Douyin's Xiaoyunque \citep{xiaoyunque} parses long scripts for short-drama production. 
Yet narrative prose is not a list of drawable entities; it often uses figurative constructions to convey imagery, affect, and abstraction \citep{mwlb,design,saakyan2025understanding}. 
In these settings, faithful generation requires deciding not only what is mentioned, but also what should become visually present.

When models make this visibility decision incorrectly, the result is a role assignment error. 
We define it as \textbf{Figurative Vehicle Intrusion}: a textually licensed vehicle that should characterize the tenor is instead instantiated as an unintended scene entity. 
Unlike hallucination, the intruding content is supported by the prompt; unlike omission, the model grounds too much text rather than too little.
However, this role-level failure has remained largely outside the scope of both TTI generation objectives and evaluation protocols. 
Recent models and alignment methods often improve image--text compatibility by encouraging generated images to follow detailed captions or preference signals more closely \citep{drawbench,dalle3,clipscore,pick,imagereward}. 
Compositional benchmarks further reward correct rendering of specified objects, attributes, counts, and relations \citep{geneval,compbench,corebench}, while QA-based protocols ask whether prompt-derived facts are recoverable from the image \citep{tifa,dsg,ihalla}. 
These criteria are effective for detecting omissions and binding errors, but they also reinforce a \textbf{\textit{presence is faithfulness}} assumption that breaks down for figurative prompts.
\begin{wrapfigure}{r}{0.62\textwidth}
    \centering
    \vspace{-0.6em}
    \includegraphics[width=0.95\linewidth]{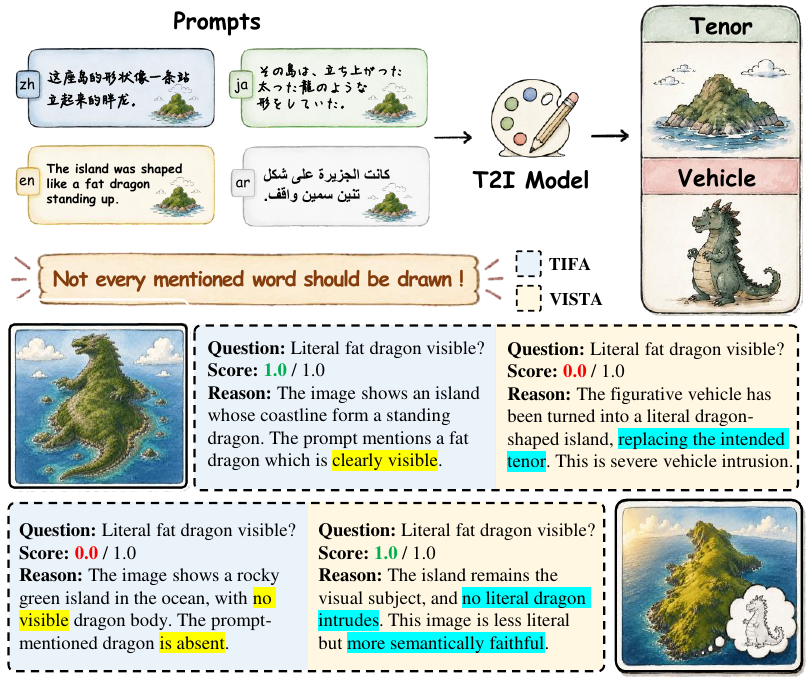}
    \caption{TIFA-style evaluation rewards vehicle intrusion.}
    \label{fig:eval-contrast}
    \vspace{-0.8em}
\end{wrapfigure}
As illustrated in Fig.~\ref{fig:eval-contrast}, in the prompt ``\textit{The island was shaped like a fat dragon standing up},'' the intended tenor is the island and the vehicle is the dragon, which means that the dragon should only describe the island's shape rather than appear as a literal object.
A TIFA-style \citep{tifa} check may nevertheless reward the generated image because the dragon is a prompt-mentioned entity that becomes visible, whereas a semantically faithful image should preserve the island as the scene subject and use the dragon only to guide its shape.
This example shows why evaluation must go beyond prompt coverage and test whether a mentioned vehicle is used as semantic guidance or rendered as an object.

It is therefore urgent to more carefully consider what TTI models should be faithful to when prompts contain figurative language: every mentioned word, or the visual scene implied by the sentence.
To make this question measurable, we build \textbf{V}ehicle \textbf{I}ntrusion and \textbf{S}emantic \textbf{T}enor \textbf{A}ssessment (\textbf{VISTA}), a benchmark composed of figurative prompts. 
Because figurative language appears across languages and varies in how comparisons are expressed and what motivates tenor--vehicle transfer, VISTA uses multilingual prompts, \textit{Figurative Form} and \textit{Mapping Mechanism} annotations, and explicit tenor--vehicle role labels.
To turn these annotations into an interpretable evaluation, we further propose \textbf{V-Score}, a diagnostic QA-based metric that evaluates tenor preservation, vehicle avoidance, scene coherence, and figurative-meaning preservation, so that a model is not rewarded merely for deleting expressive language. Using VISTA and V-Score, we evaluate five contemporary TTI models and show that vehicle intrusion persists across languages and figurative categories, while comparison with presence-oriented evaluation confirms that conventional prompt-coverage protocols can miss this failure. Finally, as a proof-of-concept mitigation, we introduce \textbf{VISTA-Guard}, a lightweight skill-based agent \citep{cc-skill} that reduces vehicle intrusion and improves overall semantic faithfulness.
To summarize, our contributions are as follows:
\begin{itemize}
\item We identify and formalize \textbf{Figurative Vehicle Intrusion}, a role-level grounding failure in figurative prompts that challenges presence-oriented assumptions about visual faithfulness.

\item We introduce \textbf{VISTA}, a multilingual benchmark organized by figurative form and semantic mapping mechanism, together with \textbf{V-Score}, providing diagnostic guidance for evaluating and improving TTI models under figurative prompts.

\item We evaluate contemporary TTI models and show that vehicle intrusion persists across languages and figurative categories, while presence-oriented evaluation can miss this failure.

\item We propose \textbf{VISTA-Guard}, a lightweight skill-based agent that significantly reduces vehicle intrusion in existing TTI models without modifying their underlying generators.
\end{itemize}

\section{Related Work}

\subsection{Advances in Text-to-Image Generation}
Modern TTI research has advanced along two coupled goals: visual fidelity and semantic control.
Imagen \citep{drawbench} improves photorealism and language understanding with large pretrained language encoders, while Parti \citep{scaling} studies autoregressive scaling for content-rich prompts.
Recent diffusion-transformer systems further improve high-resolution synthesis, text comprehension, and typography \citep{esser2024sd3}; DALL-E-3 \citep{dalle3} uses synthetic recaptioning to follow detailed descriptions more reliably.
As image quality has improved, work has increasingly targeted compositional correctness: Structured Diffusion \citep{feng2023structured} reduces missing objects and attribute-binding errors, whereas RPG \citep{yang2024rpg} plans region-wise generation for multiple objects and relations.
These advances largely treat fidelity as accurately rendering specified content.
We instead ask whether a mentioned concept should be rendered at all, since figurative prompts may use it as semantic guidance rather than as a drawable object.

\subsection{Text-to-Image Evaluation Benchmarks and Metrics}
TTI evaluation has evolved from broad capability coverage to finer-grained diagnosis.
Prompt suites such as DrawBench \citep{drawbench} test general prompt following, while holistic frameworks like HEIM \citep{HEIM} and Gecko \citep{gecko} cover alignment, aesthetics, reasoning, multilinguality, and safety-related criteria.
Compositional benchmarks \citep{geneval,corebench} instead ask whether specified objects, attributes, counts, and relations are rendered correctly.
Scalar metrics as CLIPScore \citep{clipscore} and PickScore \citep{pick} provide another view by estimating image--text compatibility or human preference.
QA-based metrics translate prompts into verifiable visual propositions: TIFA \citep{tifa} evaluates prompt-derived question--answer pairs, DSG \citep{dsg} organizes atomic questions into dependency graphs, and I-HallA \citep{ihalla} extends this paradigm to factual image hallucination.
These evaluations mainly reward the presence of prompt-derived content, whereas VISTA asks whether each concept should be visually instantiated according to its semantic role.

\subsection{Figurative Language in Visual Generation}
Figurative language conveys meaning through relations between target and source concepts rather than literal reference alone \citep{mwlb}. 
Recent work shows that common TTI captions underrepresent subjective and figurative language, limiting the coverage of nonliteral prompting in standard training data \citep{kundu2025looking,saakyan2025understanding}. 
Complementary multimodal benchmarks study literal--figurative distinctions, visual entailment, and visual metaphor understanding \citep{zhang2021multimet,yarom2023you,kleinlein2022language}. 
On the generation side, prior approaches translate nonliteral text into richer visual descriptions \citep{shahmohammadi2023vipe,chakrabarty2023spy} or introduce explicit metaphor mappings and binding mechanisms to guide synthesis \citep{huang2026cmig}. 
VISTA instead evaluates whether a figurative vehicle is assigned the wrong visual role, and VISTA-Guard provides a lightweight intervention for this role-aware setting.

\section{Methodology}

\subsection{Figurative Vehicle Intrusion}
Modern TTI models are often optimized to render prompt-mentioned entities as visible content.
However, this presence-oriented assumption does not hold for figurative descriptions such as those illustrated in Fig.~\ref{fig:eval-contrast}.
Linguistically, a figurative expression relates a \textbf{\textit{Tenor}} to a \textbf{\textit{Vehicle}} \citep{mwlb}: the tenor refers to the entity that exists in the described scene and should be visualized, whereas the vehicle does not denote an independent scene entity but provides a concept for characterizing the tenor.
When a model violates this role distinction by rendering the vehicle as a literal entity, we refer to the resulting error as \textbf{Figurative Vehicle Intrusion (FVI)}. 

Formally, let $\mathcal{C}_x$ denote the concepts mentioned in prompt $x$. 
For any concept $c$, let $O_x(c)$ indicate that $c$ should exist as an independent visual object in the intended scene, and let $G_I(c)$ indicate that $c$ is recognizably instantiated in the generated image $I$.
Presence-oriented evaluation typically requires only $\forall c\in\mathcal{C}_x,\,G_I(c)$, whereas FVI specifies the condition:
\begin{equation}
    \operatorname{FVI}(x,I)
    \Longleftrightarrow
    \exists c\in\mathcal{C}_x:
    \neg O_x(c) \;\land\; G_I(c).
\end{equation}
Thus, FVI occurs when a vehicle concept that should function only as figurative guidance is granted object-level presence in the generated image.

\begin{figure*}[t]
    \centering
    \includegraphics[width=0.98\textwidth]{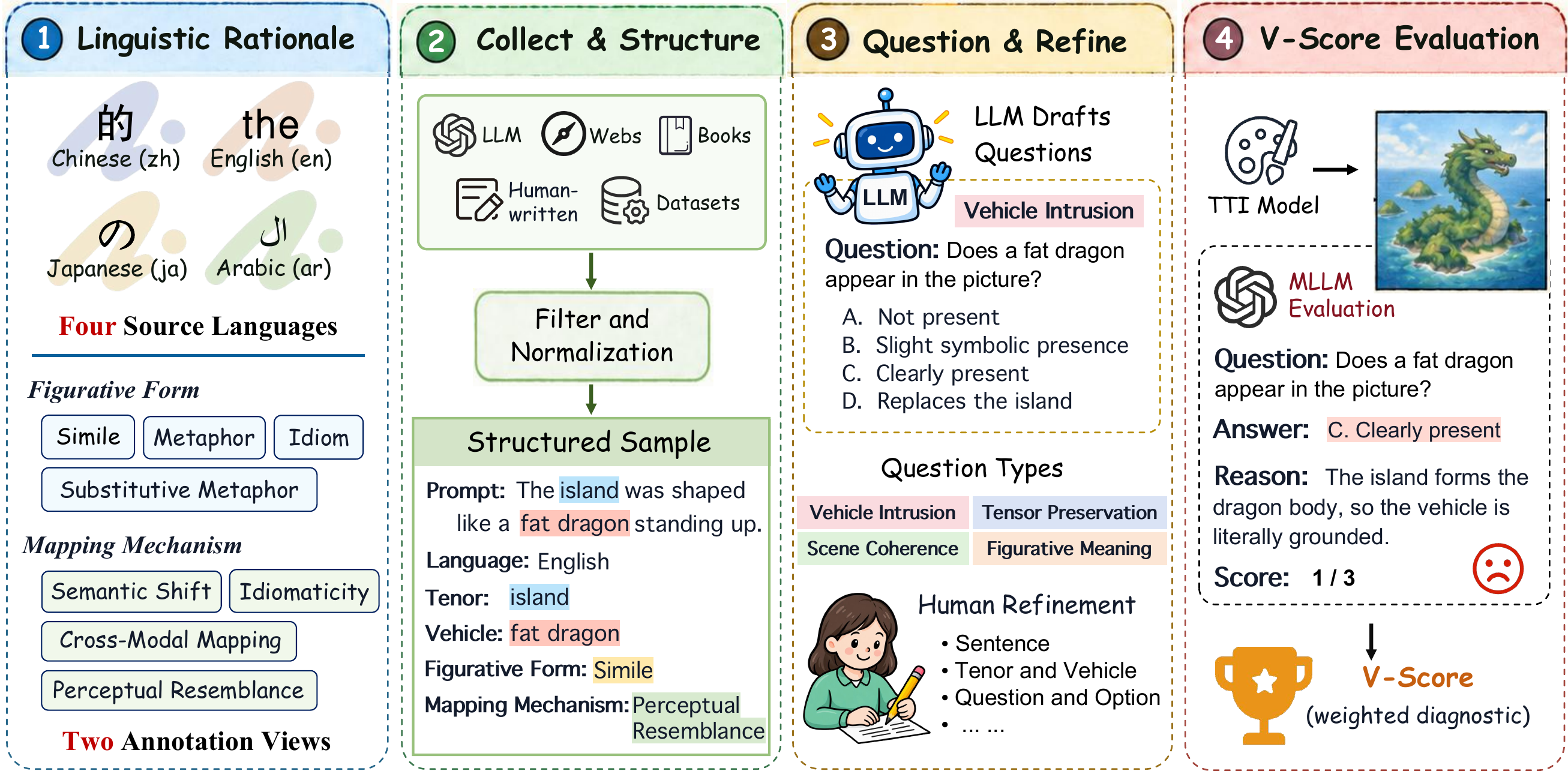}
    \caption{Overview of VISTA Benchmark Construction and Evaluation.}
    \label{fig:vista-pipeline}
    \vspace{-0.8em}
\end{figure*}

\subsection{VISTA: Benchmark for Vehicle Intrusion and Semantic Tenor Assessment}
Evaluating FVI requires prompts where figurative vehicles are separated from concepts that should appear as visual objects, a role distinction not targeted by existing TTI benchmarks.
To support this evaluation, we introduce VISTA and its QA-based V-Score, as illustrated in Fig.~\ref{fig:vista-pipeline}.

\subsubsection{Dataset Construction}

\paragraph{Linguistic Rationale.}
Figurative language appears across languages \citep{wals} and is realized through different linguistic forms and tenor--vehicle mappings \citep{mwlb}.
Accordingly, VISTA organizes the benchmark by source language and annotates each prompt from two complementary views, as shown in panel 1 of Fig.~\ref{fig:vista-pipeline}.

For language coverage, we select \textit{Chinese}, \textit{English}, \textit{Japanese}, and \textit{Arabic}, spanning Sinitic, Germanic, Japonic, and Semitic languages and covering typological differences in script, segmentation, morphology, and writing direction \citep{wals,harald_hammarstrom_2026_18840935}.
This selection introduces typological variation while retaining the same core semantic problem \citep{ploeger2024typological}, and extends evaluation beyond English-centered prompts \citep{joshi2020state}.

For annotation, VISTA uses two complementary views of the tenor--vehicle relation without reducing figurative language to a single taxonomy \citep{mwlb}.
\textbf{Figurative Form} follows metaphor-identification and multiword-expression work by recording how the relation is linguistically expressed; 
its four labels are \textit{Simile}, \textit{Metaphor}, \textit{Substitutive Metaphor}, and \textit{Idiom}.
\textbf{Mapping Mechanism} records what motivates the transfer from vehicle to tenor;
its four labels are \textit{Semantic Shift}, \textit{Perceptual Resemblance}, \textit{Cross-modal Mapping}, and \textit{Idiomaticity}.
Detailed definitions and decision boundaries for the eight labels are provided in Appendix~\ref{app:vista-taxonomy}.

\paragraph{Data Collection and Construction.}
We collect figurative candidates from five sources: LLM-assisted generation, websites, books, human-written cases, and existing datasets. 
As shown in panel 2 of Fig.~\ref{fig:vista-pipeline}, all candidates are filtered and normalized, retaining only prompts with identifiable 
\begin{wraptable}{l}{0.65\textwidth}
\centering
\vspace{-1.2em}
\caption{VISTA sample counts by language and annotation view.}
\label{tab:vista-composition}
\begin{tabular}{@{}lrrrrr@{}}
\toprule
\textbf{Annotation} & \textbf{zh} & \textbf{en} & \textbf{ja} & \textbf{ar} & \textbf{Total} \\
\midrule
\quad Simile                  & 130 & 220 & 190 & 120 & 660 \\
\quad Metaphor                & 120 & 110 & 230 & 120 & 580 \\
\quad Substitutive Metaphor   & 120 &  90 &  20 & 160 & 390 \\
\quad Idiom                   &  70 &  70 &  60 & 100 & 300 \\
\midrule
\quad Semantic Shift          & 180 & 320 & 270 & 200 & 970 \\
\quad Perceptual Resemblance  & 100 &  60 & 110 & 100 & 370 \\
\quad Cross-modal Mapping     &  90 &  40 &  60 & 100 & 290 \\
\quad Idiomaticity            &  70 &  70 &  60 & 100 & 300 \\
\midrule
\textbf{Language Total} & \textbf{440} & \textbf{490} & \textbf{500} & \textbf{500} & \textbf{1930} \\
\bottomrule
\end{tabular}
\end{wraptable}
tenor--vehicle roles and no literal obligation for the vehicle to appear.
Each sample records the prompt $x_i$, language $\ell_i$, tenor $t_i$, vehicle $v_i$, Figurative Form $f_i$, and Mapping Mechanism $m_i$:
\begin{equation}
b_i=(x_i,\ell_i,t_i,v_i,f_i,m_i).
\end{equation}
Table~\ref{tab:vista-composition} reports the sample distribution in VISTA, which preserves natural variation from language-specific conventions and source availability.

\paragraph{Question Generation}
Because FVI concerns both role assignment and overall semantic faithfulness, we generate four multiple-choice questions for each sample along two diagnostic axes, as shown in panel 3 of Fig.~\ref{fig:vista-pipeline}. 
The role-level questions assess whether the intended tenor is preserved and whether the vehicle avoids object-level realization, while the semantic-level questions assess scene coherence and preservation of the figurative meaning. 
Given the prompt and its tenor--vehicle annotations, an LLM drafts each question and its ordered response options in the source language, so that the QA items test semantic-role-aware grounding rather than generic object presence.

\paragraph{Human Refinement}
To make the generated QA sets reliable as visual evaluation items, human annotators review the original sentence, the tenor--vehicle labels, and the question--option set together. 
They verify that each question matches the intended diagnostic axis, can be answered from the image, and has mutually exclusive ordered options, revising ambiguous wording and incorrect role assignments. 
Annotators also check whether the source-language wording preserves the original figurative relation and whether the ordered options reflect meaningful severity levels rather than stylistic variants.
The resulting verified QA set is used for V-Score evaluation.

\subsubsection{V-Score: An Evaluation Metric Using Question-Answering}
Given the constructed benchmark items and their human-verified QA sets, we define V-Score to evaluate whether a generated image grounds a figurative prompt according to semantic roles rather than prompt coverage alone.
It follows the interpretable QA paradigm used in TTI evaluation \citep{tifa,li2024evaluating}, but adapts it to role-aware figurative faithfulness.
As shown in panel 4 of Fig.~\ref{fig:vista-pipeline}, for each benchmark prompt $p$, a TTI model produces an image $I^p$, and an MLLM evaluator selects the option best supported by visible evidence while returning a short rationale.

Each sample contains four ordered multiple-choice questions.
Two questions diagnose role assignment: \textit{Tenor Preservation} checks whether the intended subject remains visible, and \textit{Vehicle Avoidance} checks whether the figurative vehicle avoids object-level realization.
The other two questions serve as semantic safeguards: \textit{Scene Coherence} checks whether the image remains plausible, and \textit{Figurative Meaning} checks whether the intended figurative interpretation is still conveyed.
For each question, options are ordered from the most to the least faithful outcome and mapped to scores in $\{3,2,1,0\}$.
To calculate V-Score, we input each generated image with its four questions and answer choices into the MLLM evaluator and map the selected options through their ordered scoring rules; the evaluator template is provided in Appendix~\ref{app:evaluator-prompt}. 
Let $\mathcal{P}$ be the set of prompts and $I^p$ the image generated for prompt $p$. 
For each $p$, let $\mathcal{Q}^p=\{(q_k^p,C_k^p,r_k^p)\}_{k=1}^{4}$ contain its questions, option sets, and scoring rules.
The score is defined as:
\begin{equation}
    \operatorname{V-Score}(\mathcal{P})
    =\frac{100}{|\mathcal{P}|}
    \sum_{p\in\mathcal{P}}
    \sum_{k=1}^{4}\alpha_k\,
    \rho\!\left(
        \operatorname{VQA}(I^p,q_k^p,C_k^p),
        r_k^p
    \right),
    \quad
    \sum_{k=1}^{4}\alpha_k=1 .
\end{equation}
Here, $\operatorname{VQA}(\cdot)$ is the evaluator's selected option and $\rho(\cdot,r_k^p)\in[0,1]$ maps it to the normalized score specified by $r_k^p$.
We set $\alpha_1=\alpha_2=0.30$ and $\alpha_3=\alpha_4=0.20$, prioritizing tenor preservation and vehicle-intrusion avoidance while retaining scene plausibility and figurative meaning. 
Beyond the aggregate score, V-Score preserves the four sub-scores and evaluator rationales, enabling analysis by language, figurative form, mapping mechanism, and error type.

\subsection{VISTA-Guard: A Skill-Based Agent for Mitigating Vehicle Intrusion}
As a proof-of-concept mitigation for FVI, we design VISTA-Guard as a lightweight skill-based agent.
Since LLMs can identify the figurative expression, tenor, and vehicle from prompts, we package this role-explicitation process as a reusable skill (Appendix~\ref{app:vista-guard-skill}), following the general paradigm of tool-using and skill-based agents \citep{schick2023toolformer,cc-skill}.
The skill is specified by a description $d$ and workflow $w$.
Given a prompt $x$, the agent first decides whether to invoke the skill, constructs an auxiliary context only when needed, and then calls the generator:
\begin{equation}
    a=R_{\phi}(x,d)\in\{0,1\},
    \qquad
    z=
    \begin{cases}
        w(x), & a=1,\\
        \varnothing, & a=0,
    \end{cases}
    \qquad
    I=\mathcal{M}_{m}\!\left(x,z\right).
\end{equation}
Here, $R_{\phi}$ is a single-pass LLM controller that decides whether the prompt matches the skill description, $z$ is the context produced by the workflow when the skill is invoked, and $\mathcal{M}_{m}$ denotes inference by the TTI model conditioned on the raw prompt and any auxiliary context.
When invoked, the workflow infers the source language, figurative expression, tenor, and vehicle from $x$ alone, then produces a guarded context that preserves the tenor as the visual subject, suppresses the vehicle as a literal entity, and conveys vehicle-derived qualities indirectly through visual attributes.
This allows VISTA-Guard to reduce vehicle intrusion with minimal integration cost across generator architectures, without retraining the underlying models.

\section{Experiments and Analysis}
\subsection{Experimental Setup}
We evaluated five text-to-image (TTI) models: Nano Banana Pro \citep{nanobananapro}, GPT-Image-2 \citep{gptimage2}, DALL-E-2 \citep{dalle2}, DALL-E-3 \citep{dalle3}, and Grok Imagine \citep{grok}. 
To control evaluation cost while retaining manually verified coverage, we select 193 samples from the full VISTA pool by $1/10$ stratified sampling according to the original language and annotation composition (Appendix~\ref{app:evaluation-subset}).
For each of the 193 VISTA samples \footnote{Sample \#178 may have triggered the DALL-E content policy, leaving 192 valid samples for them.}, every model generated one image from the original sentence using the same source-language instruction template; the complete templates are provided in Appendix~\ref{app:generation-prompts}. 
We evaluated each generated image with GPT-5.5 \citep{gpt55} by presenting the four multiple-choice questions in the sample's source language. 
The evaluator was instructed to rely only on visible evidence and to select the more conservative option when the image was ambiguous. 
The four questions measure Tenor Preservation (Q1), Vehicle Intrusion (Q2), Scene Coherence (Q3), and Figurative Meaning (Q4); their ordered options are mapped to scores from 3 to 0 and aggregated into V-Score. 

\begin{figure}[htbp]
\centering
\includegraphics[width=0.98\textwidth]{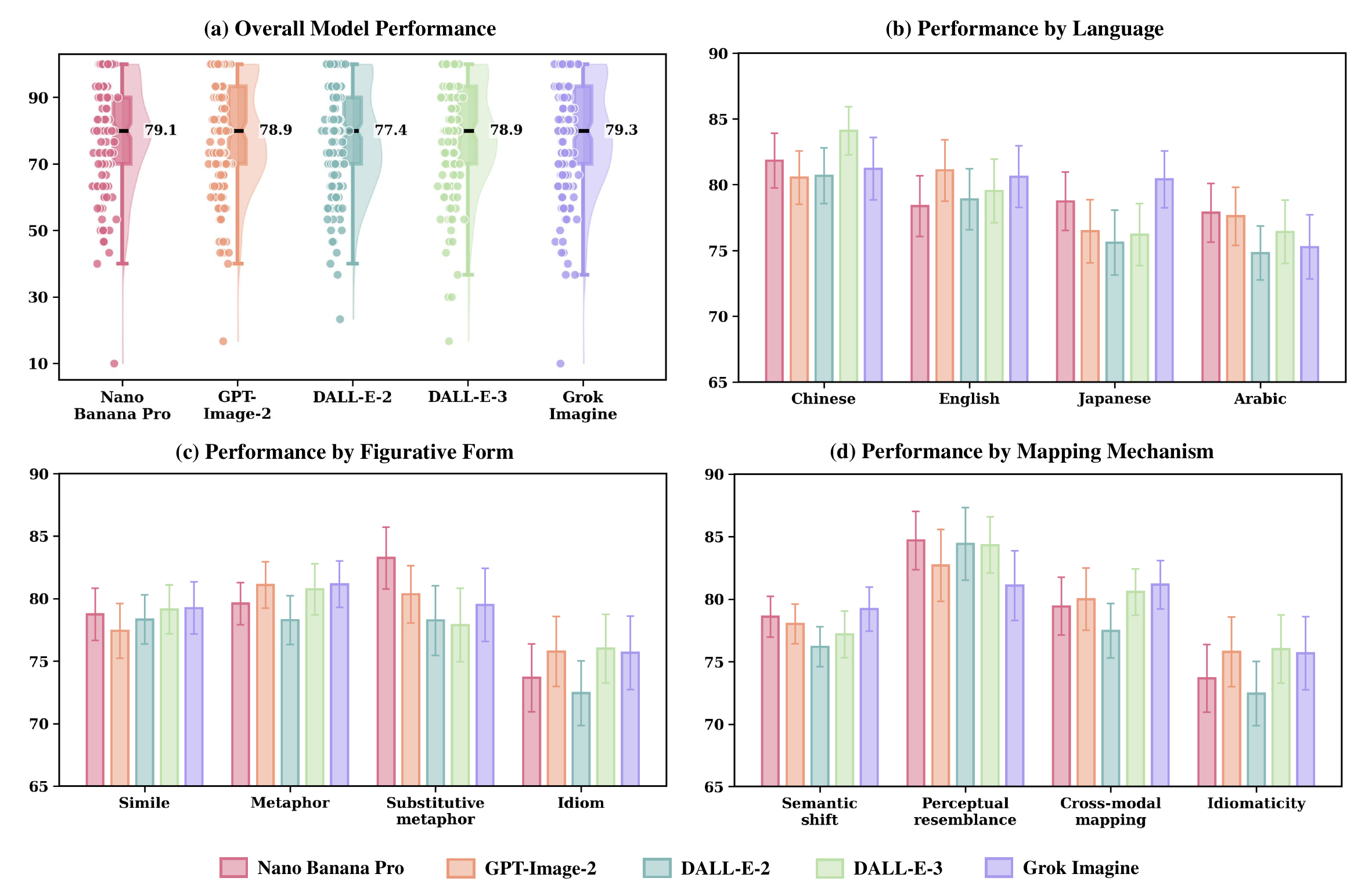}
\caption{Base performance on VISTA.
(a) Sample-level V-Score distributions for each model.
(b--d) Mean V-Score with standard errors across Languages, Figurative Forms, and Mapping Mechanisms.}
\label{fig:base-breakdown}
\end{figure}

\subsection{Evaluation of Text-to-Image Models}
Following the V-Score setting defined above, we first evaluate the five TTI models with the default weights $\alpha_1=\alpha_2=0.30$ and $\alpha_3=\alpha_4=0.20$; a weight sensitivity analysis is reported in Appendix~\ref{app:vscore-sensitivity}.
As shown in Fig.~\ref{fig:base-breakdown}(a), mean V-Scores occupy a narrow range, from 77.38 for DALL-E-2 to 79.31 for Grok Imagine, but the sample-level distributions remain broad. 
This indicates that similar aggregate scores mask substantial prompt-level variation rather than uniformly reliable behavior.
Table~\ref{tab:base-option-distribution} identifies the main source of this instability: more than 70\% of outputs receive the highest score on Q1, Q3, and Q4, but only 33.6\% do so on Q2, and 57.9\% receive 0 or 1. 
Thus, all evaluated models preserve subjects, scene coherence, and figurative meaning more reliably than they avoid literalizing the vehicle, establishing FVI as a recurrent cross-model failure.

\begin{table*}[t]
\centering
\small
\setlength{\tabcolsep}{2.5pt}
\caption{Option-score distributions for each model on Base generation. Values are percentages of valid samples assigned to scores 3--0 for each diagnostic question.}
\label{tab:base-option-distribution}
\begin{tabular*}{\textwidth}{@{\extracolsep{\fill}}ccccc|cccc|cccc|cccc@{}}
\toprule
\multirow{2}{*}{\textbf{Model}}
& \multicolumn{4}{c}{\textbf{Tenor Preservation}}
& \multicolumn{4}{c}{\textbf{Vehicle Avoidance}}
& \multicolumn{4}{c}{\textbf{Scene Coherence}}
& \multicolumn{4}{c}{\textbf{Figurative Meaning}} \\
\cmidrule(lr){2-5}\cmidrule(lr){6-9}\cmidrule(lr){10-13}\cmidrule(lr){14-17}
& \textbf{3} & \textbf{2} & \textbf{1} & \textbf{0}
& \textbf{3} & \textbf{2} & \textbf{1} & \textbf{0}
& \textbf{3} & \textbf{2} & \textbf{1} & \textbf{0}
& \textbf{3} & \textbf{2} & \textbf{1} & \textbf{0} \\
\midrule
Nano Banana Pro & 72.5 & 23.3 & 4.1 & 0.0 & 31.6 & 11.4 & 41.5 & 15.5 & 75.1 & 19.7 & 4.7 & 0.5 & 79.3 & 18.1 & 2.1 & 0.5 \\
GPT-Image-2     & 76.7 & 19.2 & 4.1 & 0.0 & 33.2 &  5.7 & 40.4 & 20.7 & 71.5 & 22.8 & 4.1 & 1.6 & 86.0 & 11.4 & 2.1 & 0.5 \\
DALL-E-2        & 69.3 & 26.6 & 4.2 & 0.0 & 32.3 &  6.8 & 37.5 & 23.4 & 69.8 & 22.9 & 7.3 & 0.0 & 81.8 & 15.6 & 2.1 & 0.5 \\
DALL-E-3        & 72.4 & 22.4 & 5.2 & 0.0 & 35.4 &  7.3 & 37.0 & 20.3 & 72.9 & 18.8 & 7.3 & 1.0 & 84.4 & 13.5 & 2.1 & 0.0 \\
Grok Imagine    & 77.2 & 18.7 & 3.6 & 0.5 & 35.8 & 10.9 & 35.8 & 17.6 & 69.9 & 22.8 & 5.7 & 1.6 & 77.7 & 17.6 & 4.1 & 0.5 \\
\bottomrule
\end{tabular*}
\end{table*}

We next stratify V-Score by Language, Figurative Form, and Mapping Mechanism in Fig.~\ref{fig:base-breakdown}(b--d). 
FVI-related errors appear in all four languages, indicating that the failure is not an English-specific artifact of prompt wording.
Arabic and Japanese are more difficult for several systems, possibly reflecting uneven multilingual coverage in image--caption data and weaker grounding for less represented scripts or linguistic structures \citep{li2023translation}. 
Across Figurative Forms, idioms are consistently difficult, plausibly because their meanings are non-compositional and cannot be recovered from literal constituents alone \citep{sag2002multiword}. 
The Mapping Mechanism view shows a related pattern: perceptual resemblance is relatively easier, likely because it offers visible correspondences such as shape or color \citep{perniss2010iconicity}, whereas idiomaticity depends more on lexicalized conventional knowledge.
Together, these results indicate that FVI is not confined to a particular language or surface form, but varies with how explicit, conventionalized, and visually recoverable the tenor--vehicle relation is.
To reduce potential evaluator bias from GPT-5.5, we further assess cross-evaluator robustness on a held-out subset in Appendix~\ref{app:cross-evaluator}.

\subsection{Comparison with Presence-Oriented Evaluation}
\begin{wraptable}{r}{0.56\textwidth}
\vspace{-2em}
\centering
\caption{TIFA-like and V-Score evaluation results.}
\label{tab:tifa-comparison}
\begin{tabular}{@{}lcccc@{}}
\toprule
\multirow{2}{*}{\textbf{Model}}
& \multicolumn{2}{c}{\textbf{TIFA-like}}
& \multicolumn{2}{c}{\textbf{V-Score}} \\
\cmidrule(lr){2-3}\cmidrule(lr){4-5}
& \textbf{Score} & \textbf{Rank} & \textbf{Score} & \textbf{Rank} \\
\midrule
Nano Banana Pro & 93.50 & 4 & 77.17 & 2 \\
GPT-Image-2     & \textbf{98.50} & \textbf{1} & 76.25 & 4 \\
DALL-E-2        & 94.50 & 2 & 73.83 & 5 \\
DALL-E-3        & 94.50 & 2 & 76.50 & 3 \\
Grok Imagine    & 92.00 & 5 & \textbf{78.50} & \textbf{1} \\
\bottomrule
\end{tabular}
\vspace{-1em}
\end{wraptable}
To determine whether conventional prompt-faithfulness evaluation exposes FVI, we compared V-Score with a TIFA-like protocol on 40 prompts sampled from the common valid set of all five models. 
The subset contains ten prompts per language and approximately preserves the distributions of Figurative Form and Mapping Mechanism.
Following TIFA \citep{tifa}, GPT-5.5 generated five source-language multiple-choice questions from each raw prompt, covering entities, attributes, relations, actions, and scene properties without access to tenor--vehicle annotations. 
Table~\ref{tab:tifa-comparison} first compares the model-level results under the two protocols. TIFA-like scores concentrate between 92.0 and 98.5, yet the resulting rankings differ markedly from those produced by V-Score. GPT-Image-2 ranks first under TIFA-like but fourth under V-Score, whereas Grok Imagine ranks fifth under TIFA-like but first under V-Score. Thus, even when evaluated on the same prompts and images, the two protocols favor different generation behavior.
We further quantified their agreement at the image level (Appendix Fig.~\ref{fig:tifa-vscore-alignment}). 
Of the 200 images, 168 (84\%) receive a TIFA-like score of 100, while their V-Scores remain widely distributed.
Correspondingly, the association between the two metrics is negligible (Spearman's $\rho=-0.042$, $N=200$).
This disagreement follows from their evaluation criteria: TIFA-like questions treat prompt-mentioned concepts as target content, so a literally rendered vehicle can satisfy the generated QA despite occupying the wrong semantic role. 
Presence-based faithfulness and V-Score therefore capture complementary properties: the former measures whether described content is recoverable, whereas the latter determines whether that content has been grounded appropriately.

\subsection{Effectiveness of VISTA-Guard}
\begin{wrapfigure}{r}{0.6\textwidth}
    \centering
    \includegraphics[width=0.98\linewidth]{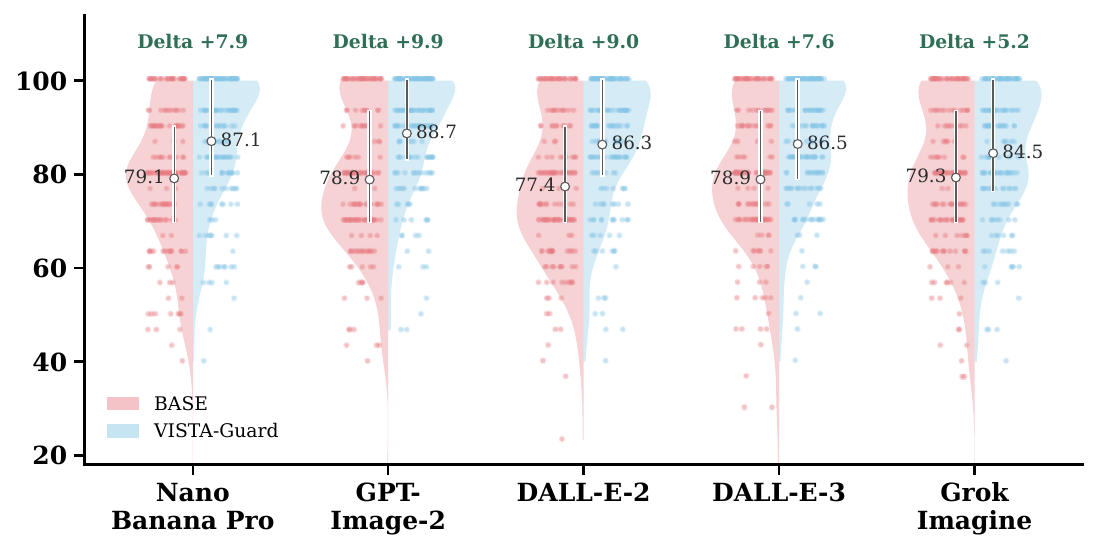}
    \caption{Sample-level V-Score distributions for Base and VISTA-Guard across five generators.}
    \label{fig:guard-distributions}
\end{wrapfigure}

To mitigate FVI, we evaluated VISTA-Guard across all five generators by pairing each guarded output with the Base image from the same prompt. As shown in Fig.~\ref{fig:guard-distributions}, VISTA-Guard consistently shifts the score distributions upward, with mean paired gains of 5.2--9.9 points.
The remaining overlap and lower-score tails indicate that it reduces, rather than eliminates, difficult failures.

We then compared VISTA-Guard with two prompt controls to determine whether its gains could be explained by prompt elaboration alone.
Human-Interact (Human) mimics an ordinary user with a minimal request to visualize the sentence, whereas Generic Anti-Hallucination (Generic Anti-Hall) adds general constraints against unsupported content without 
identifying tenor--vehicle roles.
\begin{table*}[!b]
\centering
\caption{V-Score under four prompting strategies.
$\Delta$ Score and $\Delta$ Q2 denote changes in the overall score and vehicle-avoidance score relative to Base, respectively.}
\label{tab:guard-results}
\begin{tabular}{@{}lcccccccc@{}}
\toprule
\multirow{2}{*}{\textbf{Model}}
& \multirow{2}{*}{\textbf{Human}}
& \multirow{2}{*}{\textbf{Base}}
& \multicolumn{3}{c}{\textbf{Generic Anti-Hall}}
& \multicolumn{3}{c}{\textbf{VISTA-Guard}} \\
\cmidrule(lr){4-6}\cmidrule(lr){7-9}
& &
& \textbf{Score} & \textbf{$\Delta$ Score} & \textbf{$\Delta$ Q2}
& \textbf{Score} & \textbf{$\Delta$ Score} & \textbf{$\Delta$ Q2} \\
\midrule
Nano Banana Pro & 78.2 & 79.1 & 79.5 & +0.42 & -0.13 & \textbf{87.1} & \textbf{+7.94} & +1.08 \\
GPT-Image-2     & 77.3 & 78.9 & 80.2 & +1.33 & +0.21 & \textbf{88.7} & \textbf{+9.86} & +1.18 \\
DALL-E-2        & 78.2 & 77.4 & 79.7 & +2.37 & -0.09 & \textbf{86.3} & \textbf{+8.96} & +1.25 \\
DALL-E-3        & 78.8 & 78.9 & 79.4 & +0.54 & -0.27 & \textbf{86.5} & \textbf{+7.57} & +1.13 \\
Grok Imagine    & 79.9 & 79.3 & 81.2 & +1.90 & +0.06 & \textbf{84.5} & \textbf{+5.18} & +0.76 \\
\midrule
\textbf{Mean}   & 78.5 & 78.7 & 80.0 & +1.31 & -0.04 & \textbf{86.6} & \textbf{+7.90} & +1.08 \\
\bottomrule
\end{tabular}
\end{table*}
As shown in Table~\ref{tab:guard-results}, Human-Interact remains close to Base, while Generic Anti-Hallucination improves the overall score by only 1.3 points on average and produces almost no change in Q2. 
In contrast, VISTA-Guard improves the overall score and Q2 by 7.90 and 1.08 points on average, respectively. 
These results suggest that explicit tenor--vehicle role assignment is a useful and low-cost direction for mitigating FVI, beyond generic caution against hallucination.
Prompt templates for all strategies are provided in Appendix~\ref{app:generation-prompts}.

\begin{wrapfigure}{r}{0.65\textwidth}
    \vspace{-2.2em}
    \centering
    \includegraphics[width=0.98\linewidth]{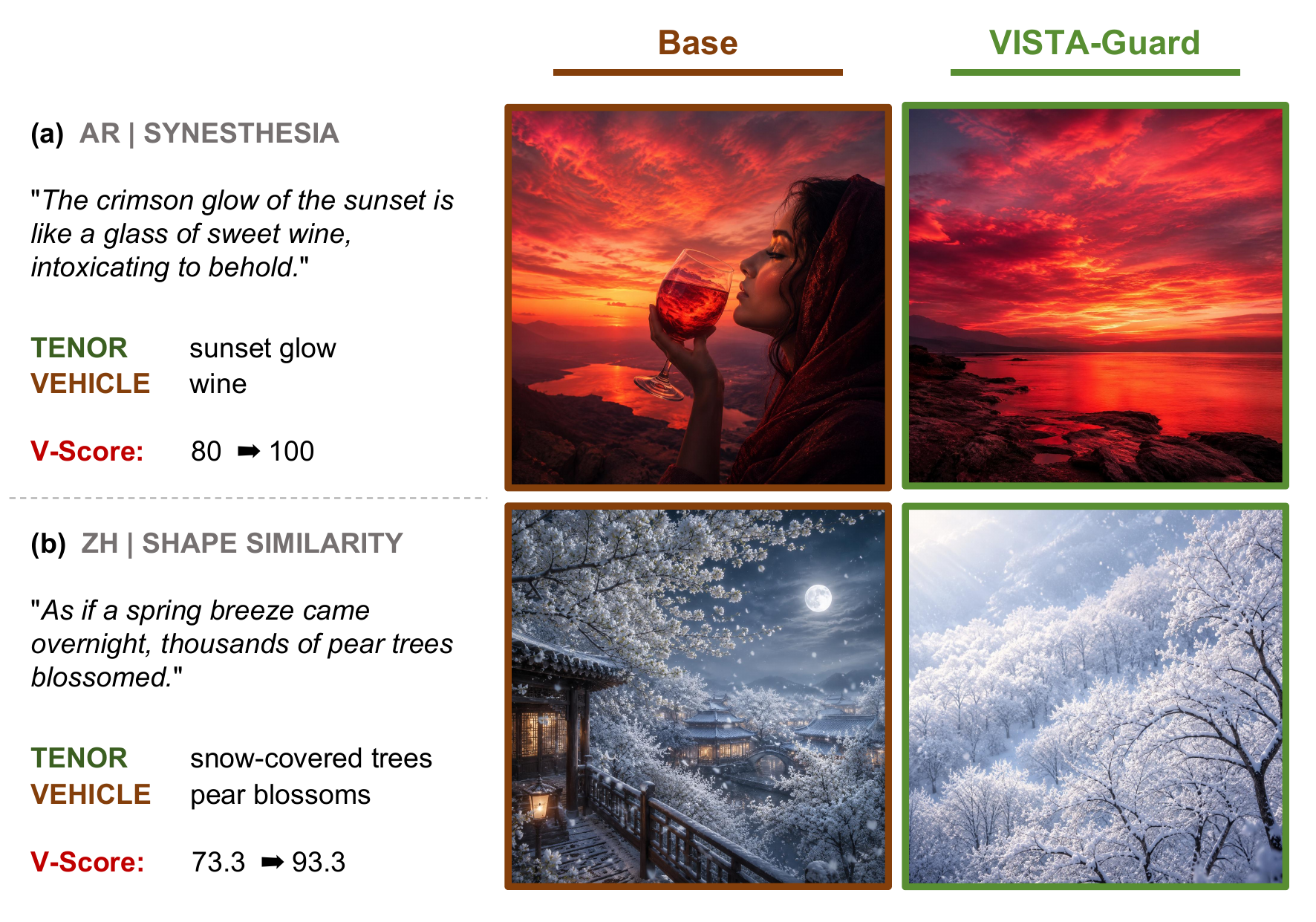}
    \caption{Two representative cases comparing Base and VISTA-Guard.
    Prompts are translated into English.}
    \label{fig:guard-case}
    % \vspace{-0.8em}
\end{wrapfigure}

To make these changes more concrete, Fig.~\ref{fig:guard-case} provides qualitative examples of the same effect. In the Arabic synesthetic simile, the intended tenor is the sunset and the vehicle is wine: Base preserves the reddish atmosphere but turns the vehicle into a foreground glass, shifting attention away from the sky. VISTA-Guard instead keeps the sunset as the scene subject and expresses the wine-like quality through crimson color and glow, raising V-Score from 80.0 to 100.0. In the Chinese shape-based metaphor, the intended tenor is snow-covered scenery and the vehicle is pear blossoms. Base literalizes the blossoms on the branches, whereas VISTA-Guard renders white snow-laden trees without adding flowers, increasing V-Score from 73.3 to 93.3.

\subsection{Reliability of V-Score through Human Evaluation}
\begin{wrapfigure}{r}{0.42\textwidth}
    \vspace{-0.8em}
    \centering
    \includegraphics[width=0.96\linewidth]{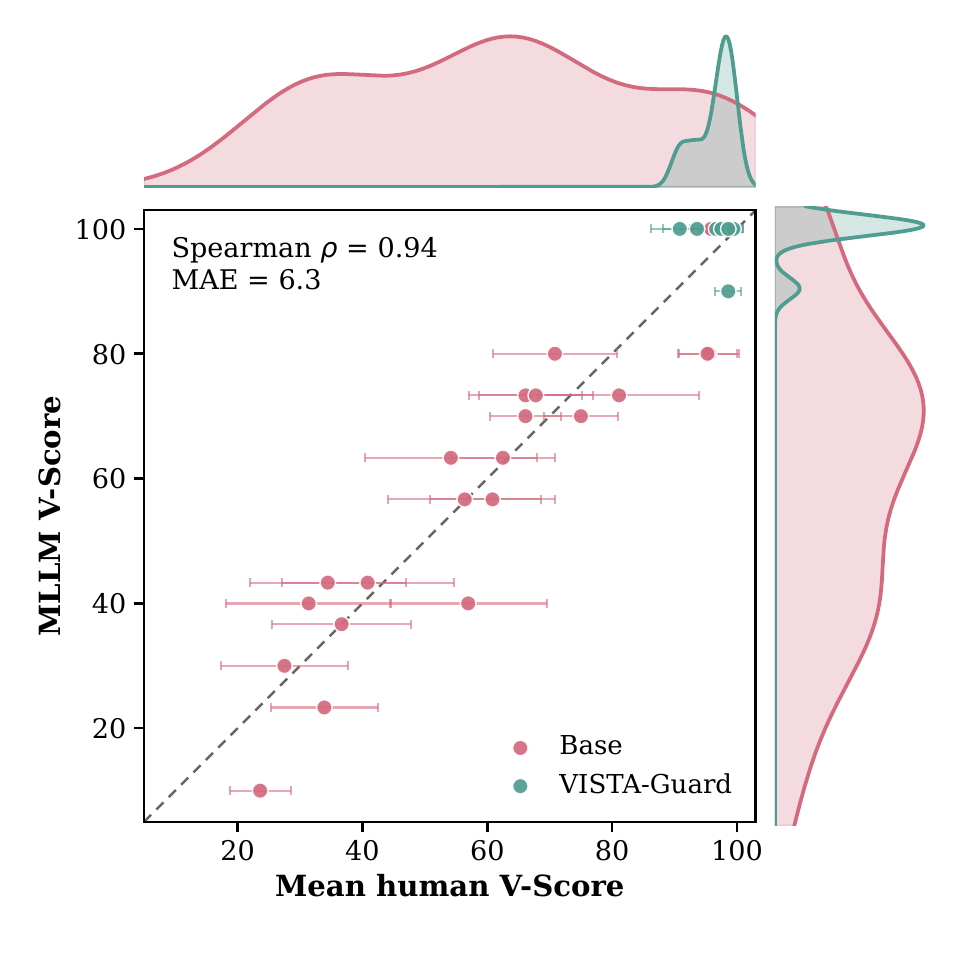}
    % \vspace{-1.5em}
    \caption{Human--MLLM agreement on 30 blind images.}
    \label{fig:human-alignment}
\end{wrapfigure}
To examine whether V-Score reflects human judgment, we recruited 12 participants to independently evaluate 30 blind images using the same four question dimensions and option-to-score mapping. 
Each image received 12 complete ratings, and human reference scores were obtained by averaging the participant scores before comparison with the MLLM evaluator.
The human ratings show strong internal reliability, with an absolute-agreement ICC of 0.784 for single raters and 0.978 for the 12-rater average.
The results are shown in Fig.~\ref{fig:human-alignment}, where each point compares the MLLM-derived V-Score with the mean human score and horizontal error bars indicate uncertainty across raters. The marginal density curves further show that the two score distributions occupy similar ranges, rather than matching only a few isolated examples. 
The points closely follow the diagonal trend from low- to high-scoring images, and the MLLM-derived V-Score correlates strongly with mean human judgment (Spearman's $\rho=0.939$, MAE = 6.3). 
The results support V-Score as a proxy for human assessment of FVI-sensitive image faithfulness.

\section{Conclusion}
This work shows that figurative prompts require TTI models to follow visual roles, not simply every mentioned word. We identify figurative vehicle intrusion, a role-level grounding failure in which a textually supported vehicle is rendered as an unintended scene entity. To make this failure measurable, VISTA and V-Score organize figurative prompts by language, Figurative Form, and Mapping Mechanism, and evaluate whether images preserve the intended tenor while avoiding literal vehicle intrusion. Experiments show that FVI persists in recent high-performing TTI models, is missed by presence-oriented evaluation, and aligns well with human judgment under our QA-based protocol. VISTA-Guard further shows that making tenor--vehicle roles explicit before generation can reduce this failure. Together, these resources provide a benchmark and practical direction for TTI models faithful not only to prompt words, but to their intended visual roles.

\section*{AI use statement}
In this work, we used generative AI tools to assist with synthetic benchmark item generation, dataset cleaning and reformatting, MLLM-based evaluation, literature search and summarization, and manuscript language polishing. AI-assisted data and evaluation outputs were manually reviewed by the authors, including checks of tenor--vehicle annotations, question--answer sets, scoring outputs, and cited literature. Generative AI tools were not used to replace author judgment in defining the research problem, validating the experimental conclusions, or determining the final claims. The authors take full responsibility for the final content of this submission, including all text, data, evaluations, and artifacts produced with the aid of generative AI.

\bibliography{iclr2027_conference}
\bibliographystyle{iclr2027_conference}

\appendix
\section{Appendix}

\subsection{Detailed VISTA Annotation Taxonomy}
\label{app:vista-taxonomy}
VISTA assigns each sample one label under \textit{Figurative Form} and one under \textit{Mapping Mechanism}. These views are complementary rather than hierarchical: the former describes how the tenor--vehicle relation is linguistically expressed, whereas the latter describes what makes the transfer interpretable \citep{mwlb,design,group2007mip}.
The main text lists the category names for readability; this appendix provides the definitions and decision boundaries used during annotation.

\subsubsection{Language Coverage}
The selected languages cover distinct families, scripts, segmentation conventions, and writing directions. This design introduces typological variation while keeping the semantic problem of tenor--vehicle role assignment comparable across languages.
\begin{table}[h]
\centering
\small
\label{tab:app-language-coverage}
\begin{tabular}{@{}lllll@{}}
\toprule
\textbf{Language} & \textbf{Family} & \textbf{Writing system} & \textbf{Segmentation} & \textbf{Direction} \\
\midrule
Chinese & Sinitic & Logographic & No whitespace word segmentation & left-to-right \\
English & Germanic & Alphabetic & Whitespace-delimited words & left-to-right \\
Japanese & Japonic & Mixed scripts & No whitespace word segmentation & left-to-right \\
Arabic & Semitic & Abjad & Morphologically rich & right-to-left \\
\bottomrule
\end{tabular}
\end{table}

\subsubsection{Figurative Form}
Figurative Form describes how the tenor--vehicle relation is expressed. The labels distinguish explicit comparison, implicit attribution, referential replacement, and lexicalized expression.
\begin{table}[h]
\centering
\small
\begin{tabular}{@{}llp{0.5\linewidth}@{}}
\toprule
\textbf{Label} & \textbf{Core distinction} & \textbf{Decision boundary} \\
\midrule
Simile & Explicit comparison & Tenor and vehicle are linked by a marker equivalent to ``like'' or ``as.'' \\
Metaphor & Implicit attribution & Vehicle characterizes an explicit or recoverable tenor without an overt comparison marker. \\
Substitutive Metaphor & Referential replacement & Vehicle expression stands in for, or partially replaces, the tenor, which must be recovered from context \citep{zhang2015exploring}. \\
Idiom & Lexicalized form & Conventionalized multiword expression conveys a non-compositional meaning \citep{sag2002multiword}. \\
\bottomrule
\end{tabular}
\end{table}

\subsubsection{Mapping Mechanism}
Mapping Mechanism describes what makes the transfer from vehicle to tenor interpretable. The labels distinguish conceptual movement, perceptual similarity, cross-modal transfer, and conventionalized meaning.
\begin{table}[h]
\centering
\small
\begin{tabular}{@{}llp{0.45\linewidth}@{}}
\toprule
\textbf{Label} & \textbf{Core distinction} & \textbf{Decision boundary} \\
\midrule
Semantic Shift & Conceptual transfer & Interpretation depends on abstraction, concretization, or related conceptual movement rather than perceptual similarity. \\
Perceptual Resemblance & Shared perceptual property & Mapping is licensed by visible or audible similarity, such as shape or sound \citep{perniss2010iconicity,cwiek2021bouba}. \\
Cross-modal Mapping & Sensory transfer & A property associated with one modality characterizes an experience in another, as in synesthetic expressions \citep{duffy2010synaesthesia}. \\
Idiomaticity & Conventional meaning & Interpretation relies on lexicalized, non-compositional knowledge rather than a productive conceptual or perceptual mapping \citep{sag2002multiword}. \\
\bottomrule
\end{tabular}
\end{table}

\subsection{Prompts}
\label{app:prompts}
This section collects the prompt templates used for image generation and evaluation.

\subsubsection{Generation Prompts}
\label{app:generation-prompts}
We use source-language-matched generation templates for all TTI models. The English template below is shown for readability; Chinese, Japanese, and Arabic samples use corresponding translations. The placeholder \texttt{[CONTENT]} is replaced by the corresponding VISTA sentence.

\begin{tcolorbox}[
    title=Human-Interact Prompt,
    colback=white,
    colframe=blue!55!black,
    colbacktitle=blue!12,
    coltitle=black,
    fonttitle=\bfseries,
    boxrule=0.6pt,
    arc=2pt,
    left=6pt,
    right=6pt,
    top=5pt,
    bottom=5pt,
    breakable
]
\small
Please generate an image based on the following sentence:\\

\texttt{[CONTENT]}
\end{tcolorbox}

\begin{tcolorbox}[
    title=Base Generation Prompt,
    colback=white,
    colframe=blue!55!black,
    colbacktitle=blue!12,
    coltitle=black,
    fonttitle=\bfseries,
    boxrule=0.6pt,
    arc=2pt,
    left=6pt,
    right=6pt,
    top=5pt,
    bottom=5pt,
    breakable
]
Generate exactly one image that directly illustrates the following sentence.\\

\textbf{Sentence:} \texttt{[CONTENT]}\\

Use the sentence's language context, cultural meaning, and figurative meaning as visual guidance. Make the image a coherent visual scene rather than a diagram or poster. Return an image only. \\

Do not explain, analyze, translate, summarize, or answer in text. Do not include any readable text, subtitles, captions, labels, signs, handwriting, letters, numbers, logos, or watermarks anywhere in the image.
\end{tcolorbox}

\begin{tcolorbox}[
    title=Generic Anti-Hallucination Prompt,
    colback=white,
    colframe=blue!55!black,
    colbacktitle=blue!12,
    coltitle=black,
    fonttitle=\bfseries,
    boxrule=0.6pt,
    arc=2pt,
    left=6pt,
    right=6pt,
    top=5pt,
    bottom=5pt,
    breakable
]
\small
Generate exactly one image that directly illustrates the following sentence.\\

\textbf{Sentence:} \texttt{[CONTENT]}\\

Use the sentence's language context, cultural meaning, and figurative meaning as visual guidance. Make the image a coherent visual scene rather than a diagram or poster. Return an image only. 
Do not explain, analyze, translate, summarize, or answer in text. Do not include any readable text, subtitles, captions, labels, signs, handwriting, letters, numbers, logos, or watermarks anywhere in the image.\\

\textbf{General anti-hallucination constraints:}\\
Only depict elements that are visually necessary for a coherent interpretation of the sentence.
Do not add unrelated concrete objects, characters, props, symbols, decorations, or background elements.
Avoid over-literalizing figurative or poetic language into extra objects unless those objects are required by the scene itself.
If a phrase is ambiguous or nonliteral, prefer a natural visual scene that preserves the intended mood and meaning without inventing additional entities.
Keep the composition focused, realistic in internal logic, and visually self-consistent.
\end{tcolorbox}

\subsubsection{Evaluator Prompt}
\label{app:evaluator-prompt}
We use a fixed MLLM evaluator prompt that asks the model to inspect only visible image evidence, answer the four multiple-choice questions, and return structured option selections with rationales.
The following English template is shown for readability; source-language-matched translations are used for Chinese, Japanese, and Arabic samples.

\begin{tcolorbox}[
    title=MLLM Evaluator Prompt,
    colback=white,
    colframe=blue!55!black,
    colbacktitle=blue!12,
    coltitle=black,
    fonttitle=\bfseries,
    boxrule=0.6pt,
    arc=2pt,
    left=6pt,
    right=6pt,
    top=5pt,
    bottom=5pt,
    breakable
]
You are a strict visual evaluator for a benchmark.\\

Your task is NOT to creatively interpret the image.
Inspect only what is visible in the image and answer the multiple-choice questions.
Do not assume something is present just because it appears in a question.
If visual evidence is ambiguous, choose the more conservative option. \\
For questions about whether a literal comparison object appears, count recognizable real objects, icons, written marks, or symbolic motifs as visible; do not count only mood, color, lighting, or composition as the object itself. \\

\textbf{Evaluation steps:} \\
1. Read the four questions. \\
2. Inspect the image carefully. \\
3. For each question, compare all four options and choose the best supported answer. \\
4. Return only minified JSON with A/B/C/D answers. \\

\textbf{Questions:} \texttt{[QUESTIONS]} \\

Return ONLY minified JSON in this format:
\begin{verbatim}
{
  "answers":    {"Q1": "A", "Q2": "B", "Q3": "A", "Q4": "C"},
  "reasoning":  {"Q1": "...", "Q2": "...", 
                 "Q3": "...", "Q4": "..."}
}
\end{verbatim}
\end{tcolorbox}

\subsection{VISTA-Guard Skill Specification}
\label{app:vista-guard-skill}

This section provides the reusable VISTA-Guard skill specification used in our implementation.

\begin{tcolorbox}[
    title=VISTA-Guard.md,
    colback=white,
    colframe=blue!55!black,
    colbacktitle=blue!12,
    coltitle=black,
    fonttitle=\bfseries,
    boxrule=0.6pt,
    arc=2pt,
    left=6pt,
    right=6pt,
    top=5pt,
    bottom=5pt,
    breakable
]
------ \\
\textbf{Name:} VISTA-Guard. \\
\textbf{Description:} Identify figurative prompts whose vehicle should guide the depiction of a tenor without becoming an independent literal object; infer the figurative expression, tenor, and vehicle; and construct a guarded generation context that suppresses literal vehicle intrusion while preserving the intended figurative effect. \\
------ \\

\textbf{Use when:} A raw image-generation prompt contains a figurative expression in which the vehicle should influence the tenor's visual qualities but should not appear as a separate scene entity. \\

\textbf{Input:} The original image-generation prompt. \\

\textbf{Workflow:}
\begin{enumerate}
    \item Inspect the raw prompt and infer its source language.
    \item Identify whether the prompt contains a figurative expression relevant to visual grounding.
    \item If invoked, infer the tenor and vehicle from the prompt alone.
    \item Preserve the tenor as the main visual subject.
    \item Suppress the vehicle as an independent literal entity.
    \item Convey vehicle-derived qualities indirectly through visual attributes such as shape, color, lighting, texture, composition, posture, atmosphere, or emotional tone.
    \item Compose a model-ready guarded context for image generation. \\
\end{enumerate} 

\textbf{Output:} A guarded generation context that preserves the original prompt, states the inferred tenor and vehicle roles, discourages literal vehicle realization, and specifies how the intended figurative qualities should be conveyed indirectly. \\

\textbf{Do not use when:} The prompt is fully literal, both compared entities are intended to appear, or the literal coexistence of tenor and vehicle is an intentional visual pun. 
\end{tcolorbox}

\subsection{Evaluation Subset Composition}
\label{app:evaluation-subset}

The main experiments use a 193-sample subset selected from the 1,930-sample VISTA pool by $1/10$ stratified sampling. 
Table~\ref{tab:app-evaluation-subset} reports the resulting composition, which preserves the language and annotation-view coverage of the full benchmark.
\begin{table}[h]
\centering
\small
\caption{Composition of the 193-sample evaluation subset.}
\label{tab:app-evaluation-subset}
\begin{tabular}{@{}lrrrrr@{}}
\toprule
\textbf{Annotation} & \textbf{zh} & \textbf{en} & \textbf{ja} & \textbf{ar} & \textbf{Total} \\
\midrule
\quad Simile                  & 13 & 22 & 19 & 12 & 66 \\
\quad Metaphor                & 12 & 11 & 23 & 12 & 58 \\
\quad Substitutive Metaphor   & 12 &  9 &  2 & 16 & 39 \\
\quad Idiom                   &  7 &  7 &  6 & 10 & 30 \\
\midrule
\quad Semantic Shift          & 18 & 32 & 27 & 20 & 97 \\
\quad Perceptual Resemblance  & 10 &  6 & 11 & 10 & 37 \\
\quad Cross-modal Mapping     &  9 &  4 &  6 & 10 & 29 \\
\quad Idiomaticity            &  7 &  7 &  6 & 10 & 30 \\
\midrule
\textbf{Language Total} & \textbf{44} & \textbf{49} & \textbf{50} & \textbf{50} & \textbf{193} \\
\bottomrule
\end{tabular}
\end{table}

\subsection{V-Score Weight Sensitivity}
\label{app:vscore-sensitivity}

To test whether the main findings depend on the default V-Score weights, we recompute scores under several aggregation schemes: role-focused scoring using Q1+Q2, semantic-safeguard scoring using Q3+Q4, equal weighting over all four questions, and the default weighting used in the main paper. As shown in Table~\ref{tab:app-vscore-sensitivity}, equal-weight scores remain close to the default V-Score results, and the two variants are strongly correlated at the image level (Spearman's $\rho=0.988$). The Q3+Q4 scores are consistently higher and resemble conventional evaluation criteria that emphasize scene plausibility and semantic recoverability, confirming that such criteria can obscure vehicle-intrusion failures.

\begin{table}[h]
\centering
\caption{V-Score sensitivity to different aggregation schemes on Base generation.}
\label{tab:app-vscore-sensitivity}
\begin{tabular}{@{}lcccc@{}}
\toprule
\textbf{Model} & \textbf{Q1+Q2} & \textbf{Q3+Q4} & \textbf{Equal} & \textbf{Default} \\
\midrule
Nano Banana Pro & 71.24 & 90.93 & 81.09 & 79.12 \\
GPT-Image-2     & 70.64 & 91.19 & 80.92 & 78.86 \\
DALL-E-2        & 68.84 & 90.19 & 79.51 & 77.38 \\
DALL-E-3        & 70.83 & 90.97 & 80.90 & 78.89 \\
Grok Imagine    & 72.88 & 88.95 & 80.92 & 79.31 \\
\bottomrule
\end{tabular}
\end{table}

\subsection{Cross-Evaluator Robustness}
\label{app:cross-evaluator}

To examine whether V-Score is sensitive to the choice of MLLM evaluator, we additionally evaluate the same 40-prompt, 200-image subset used in the metric-comparison experiment with \texttt{qwen3.8-max}. As shown in Table~\ref{tab:app-cross-evaluator}, the second evaluator produces slightly lower absolute scores, but largely preserves the ranking pattern and the conclusion that FVI persists across generators.

\begin{table}[h]
\centering
\small
\caption{Cross-evaluator robustness of V-Score on the 40-prompt subset.}
\label{tab:app-cross-evaluator}
\begin{tabular}{@{}lcccc@{}}
\toprule
\multirow{2}{*}{\textbf{Model}}
& \multicolumn{2}{c}{\textbf{Main Evaluator}}
& \multicolumn{2}{c}{\textbf{Qwen Evaluator}} \\
\cmidrule(lr){2-3}\cmidrule(lr){4-5}
& \textbf{Score} & \textbf{Rank} & \textbf{Score} & \textbf{Rank} \\
\midrule
Nano Banana Pro & 77.17 & 2 & 75.72 & 2 \\
GPT-Image-2     & 76.25 & 4 & 74.32 & 3 \\
DALL-E-2        & 73.83 & 5 & 71.15 & 5 \\
DALL-E-3        & 76.50 & 3 & 74.10 & 4 \\
Grok Imagine    & 78.50 & 1 & 76.35 & 1 \\
\bottomrule
\end{tabular}
\end{table}

\subsection{Supplementary Metric Comparison}
\label{app:metric-comparison}

\begin{figure}[h]
    \centering
    \includegraphics[width=0.4\linewidth]{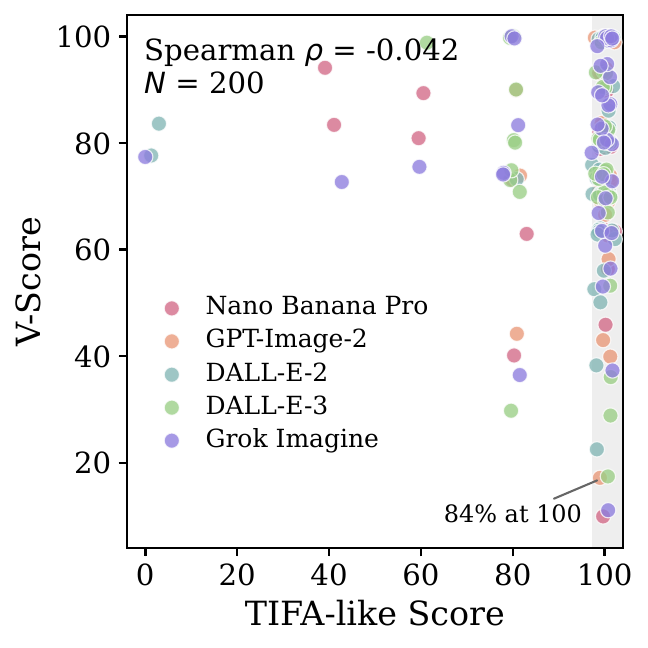}
    \caption{Image-level relationship between TIFA-like score and V-Score on 200 images. Points are slightly jittered for visibility.}
    \label{fig:tifa-vscore-alignment}
\end{figure}

Figure~\ref{fig:tifa-vscore-alignment} provides the comparison between the TIFA-like protocol and V-Score. 
Although many images receive saturated TIFA-like scores, their V-Scores remain broadly distributed, illustrating that prompt-coverage evaluation and role-aware figurative evaluation capture different failure modes.

\end{document}

%% file: math_commands.tex
\usepackage{amsmath,amsfonts,bm}

\def\eqref#1{equation~\ref{#1}}
\def\1{\bm{1}}

\DeclareMathAlphabet{\mathsfit}{\encodingdefault}{\sfdefault}{m}{sl}
\SetMathAlphabet{\mathsfit}{bold}{\encodingdefault}{\sfdefault}{bx}{n}